\documentclass{llncs}
\usepackage{makeidx}  
\usepackage{graphicx}
\graphicspath{{img/}}
\usepackage{float}
\usepackage{amsmath}
\usepackage{amssymb}

\usepackage{subcaption}
\usepackage{multirow}
\usepackage{array}
\newcolumntype{L}[1]{>{\raggedright\let\newline\\\arraybackslash\hspace{0pt}}m{#1}}
\newcolumntype{C}[1]{>{\centering\let\newline\\\arraybackslash\hspace{0pt}}m{#1}}
\newcolumntype{R}[1]{>{\raggedleft\let\newline\\\arraybackslash\hspace{0pt}}m{#1}}
\usepackage{algorithm}
\usepackage[noend]{algpseudocode}
\usepackage[misc]{ifsym}

\usepackage{alphalph}
\usepackage{etoolbox}

\makeatletter
\patchcmd{\subequations}
  {\alph{equation}}
  {-\two@digits{\arabic{equation}}}
  {}{}
\makeatother

\usepackage{float}
\floatstyle{ruled}
\newfloat{model}{thp}{lol}
\floatname{model}{Model}

\allowdisplaybreaks
\floatstyle{boxed}
\newfloat{listing}{tb}{lol}
\floatname{listing}{Listing}

\begin{document}
\mainmatter              
\title{{\large Constraint and Mathematical Programming Models} \\ {\large for Integrated Port Container Terminal Operations}}
\titlerunning{Integrated Port Container Terminal Operations}  
%
\author{Damla Kizilay\inst{1,2} \and Deniz T. Eliiyi\inst{2} \and Pascal Van Hentenryck\inst{1}}
\authorrunning{Kizilay, Eliiyi, and Van Hentenryck} 
%
\tocauthor{Damla}
\institute{\textsuperscript{1}University of Michigan, Ann Arbor MI 48109, USA\\
           \textsuperscript{2}Yasar University, Izmir 35100, Turkey}

\maketitle              

\begin{abstract}
  This paper considers the integrated problem of quay crane
  assignment, quay crane scheduling, yard location assignment, and
  vehicle dispatching operations at a container terminal. The main
  objective is to minimize vessel turnover times and maximize the
  terminal throughput, which are key economic drivers in terminal
  operations. Due to their computational complexities, these problems
  are not optimized jointly in existing work. This paper revisits this
  limitation and proposes Mixed Integer Programming (MIP) and
  Constraint Programming (CP) models for the integrated problem, under
  some realistic assumptions. Experimental results show that the MIP
  formulation can only solve small instances, while the CP model finds
  optimal solutions in reasonable times for realistic instances
  derived from actual container terminal operations.
\end{abstract}


\section{Introduction}

Maritime transportation has significant benefits in terms of cost and
capability for carrying a higher number of cargos. Indeed, sea trade
statistics indicate that 90\% of global trade is performed by maritime
transportation. This has led to new investments in container terminals
and a variety of initiatives to improve the operational efficiency of
existing terminals. Operations at a container terminal can be
classified as quay side and yard side. They handle materials using
quay cranes (QC), yard cranes (YC), and transportation vehicles such
as yard trucks (YT). QCs load and unload containers at the quay side,
while YCs load and discharge containers at the yard side. YTs provide
transshipment of the containers between the quay and the yard sides.

In a typical container terminal, it is important to minimize the
vessel berthing time, i.e., the period between the arrival and the
departure of a vessel. When a vessel arrives at the terminal, the
berth allocation problem selects when and where in the port the vessel
should berth. Once a vessel is berthed, its stowage plan determines
the containers to be loaded/discharged onto/from the vessel. This
provides an input to the {\em QC assignment and scheduling}, which
determines the sequence of the containers to be loaded or discharged
from different parts of the vessel by the QCs. In addition, the
containers discharged by the QCs are placed onto YTs and transported
to the storage area, which corresponds to a {\em vehicle dispatching
  problem}. Each discharged container is assigned a storage location,
giving rise to a {\em yard location assignment problem}. Finally, the
containers are taken from YTs by YCs and placed onto stacks in storage
blocks, specifying a {\em YC assignment and scheduling problem}.

A container terminal aims at completing the operations of each berthed
vessel as quickly as possible to minimize vessel waiting times at the
port and thus to maximize the turnover, i.e., the number of handled
containers. Optimizing the integrated operations within a containing
terminal is computationally challenging \cite{Vis2003}. Therefore, the
optimization problems identified earlier are generally
considered separately in the literature, and the number of studies
considering integrated operations is rather limited. However, although
the optimization of individual problems brings some operational
improvements, the main opportunity lies in optimizing terminal
operations holistically.  This is especially important since the
optimization sub-problems have conflicting objectives that can
adversely affect the overall performance of the system.

This paper considers the integrated optimization of container terminal
operations and proposes MIP and CP formulations under some realistic
assumptions. To the best of our knowledge, the resulting optimization
problem has not been considered in the literature so far.
Experimental results show that the MIP formulation is not capable of
solving instances of practical relevance, while the CP model finds
optimal solutions in reasonable times for realistic instances derived
from real container terminal operations.

The rest of the paper is organized as follows. Section
\ref{section-problem} specifies the problem and the
assumptions considered in this work. Section
\ref{section-related-work} provides a detailed literature review for
the integrated optimization of container terminal operations. Section
\ref{section-MIP} presents the MIP model, while Section
\ref{section-CP} presents the constraint programming model for the
same problem. Section \ref{section-experiments} presents the data
generation procedure, the experimental results, and the comparison of
the different models. Finally, Section \ref{section-conclusion}
presents concluding remarks and future research directions.

\section{Problem Definition}
\label{section-problem}

This section specifies the Integrated Port Container Terminal Problem
(IPCTP) and its underlying assumptions. The IPCTP is motivated by the
operations of actual container terminals in Turkey.

In container terminals, berth allocation assigns a berth and a time
interval to each vessel. Literature surveys and interviews with port
management officials reveal that significant factors in berth
allocation include priorities between customers, berthing privileges
of certain vessels in specific ports, vessel sizes, and depth of the
water. Because of all these restrictions, the number of alternative
berth assignments is quite low, especially in small Turkish ports. As
a result, the berthing plan is often determined without the need of
intelligent systems and the berthing decisions can be considered as
input data to the scheduling of the material-handling equipment.  The
vessel stowage plan decides how to place the outbound containers on
the vessel and is prepared by the shipping company.  These two
problems are thus separated from the IPCTP. In other words, the paper
assumes that vessels are already berthed and ready to be served.

The IPCTP is formulated by considering container groups, called {\em
  shipments}. A single shipment represents a group of containers that
travel together and belong to the same customer. Therefore, the
containers in a single shipment must be stored in the same yard block
and in the same vessel bay. In addition, each shipment is handled as a
single batch by QCs and YCs.

The IPCTP determines the storage location in the yard for inbound
containers. The yard is assumed to be divided into a number of areas
containing the storage blocks. Each inbound shipment has a number of
possible location points in each area.  Each YC is assumed to be
dedicated to a specific area of the yard. Note that outbound shipments
are at specified yard locations at the beginning of the planning
period and hence their YCs are known in advance. In contrast, for
inbound shipments, the YC assignment is derived from the storage block
decisions. The IPCTP assumes that each yard location can store at most
one shipment but there is no difficulty in relaxing that assumption.

The inbound and outbound shipments and their vessel bays are specified
in the vessel stowage plan. The IPCTP assigns each shipment to a QC
and schedules each QC to process a sequence of shipments. The QC
scheduling is constrained by movement restrictions and safety
distances between QCs. Two adjacent QCs must be apart from each other
by safety distance, so that they can perform their tasks
simultaneously without interference as described in \cite{NAV:NAV20121}.

The IPCTP assumes the existence of a sufficient number of YTs so that
cranes never wait. This assumption is motivated by observations in
real terminals where many YTs are dedicated to each QC in order to
ensure a smooth operation. This organization is justified by the fact
that QCs are the most critical handling equipment in the terminal. As
a result, QCs are very rarely blocked while discharging and almost
never starve while loading. Therefore, the assumption of having a
sufficient number of YTs is realistic and simplifies the IPCTP.

\begin{figure}[!h]
	\centering
	\includegraphics[width=0.8\linewidth]{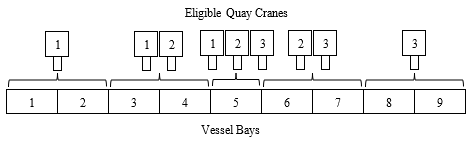}
	\caption{The Vessel Bays and Their Available QCs.}
	\label{figure-1}
\end{figure}

The IPCTP also assumes that the handling equipment (QC, YC,
YT) is homogeneous, and their processing times are deterministic and
known. Since the QCs cannot travel beyond the berthed vessel bays and
must obey a safety distance \cite{Sammarra2007}, each shipment can
only be assigned to an eligible set of QCs that respect safety
distance and non-crossing constraints. These are illustrated in Figure
\ref{figure-1} where berthed vessel bays and QCs are indexed in
increasing order and the safety distance is assumed to be 1 bay. For
instance, only QC-1 is eligible to service bays 1--2.  Similarly, only
QC-3 can operate on vessel bays 8--9. In contrast, bays 3--4 can be
served by QC-1 and QC-2.

The main objective of a container terminal is to maximize total profit
by increasing productivity. Terminal operators try to lower vessel
turn times and decrease dwell times. To lower vessel turn times, the
crane operations must be well-coordinated and the storage location of
the inbound shipments must be chosen carefully, since they impact the
distance traveled by the YTs. Therefore, the IPCTP jointly considers
the storage location assignment for the inbound shipments from
multiple berthed vessels and the crane assignment and scheduling for
both outbound and inbound containers. The objective of the problem is to
minimize the sum of weighted completion times of the vessels.

\begin{figure}[!t]
	\centering
	\includegraphics[width=0.8\linewidth]{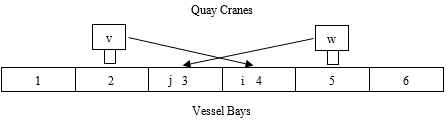}
	\caption{An Example of Interference for Shipments $i, j$ and Quay Cranes $v, w$.}
	\label{figure-2}
\end{figure}

The input parameters of the IPCTP are given in Table
\ref{table-parameter}.  Most are self-explanatory but some necessitate
additional explanation. The smallest distance $\delta_{v,w}$
between quay cranes $v$ and $w$ is given by
$\delta_{v,w}=\left(\delta+1\right)\left|v-w\right|$ where $\delta$ is
the safety distance. The minimum time between the starting times of
shipments $i$ and $j$ when processed by cranes $v$ and $w$ is given by
\[
    \Delta_{i,j}^{v,w}=
\begin{cases}
    \left(b_{i}-b_{j}+\delta_{v,w}\right)s_{QC}& \mbox{if } v < w \mbox{ and } i \ne j \mbox{ and } b_{i}>b_{j}-\delta_{v,w}\\
    \left(b_{j}-b_{i}+\delta_{v,w}\right)s_{QC}& \mbox{if } v > w \mbox{ and } i \ne j \mbox{ and } b_{i}<b_{j}-\delta_{v,w}\\
    0              & \mbox{otherwise.}
\end{cases}
\]
This captures the time needed for a quay crane to travel to a safe
distance in case of potential interference. This is illustrated in
Figure \ref{figure-2}. If shipments $i$ and $j$ are processed by
cranes $v$ and $w$, then their starting times must be separated by
$s_{QC}$ time units in order to respect the safety constraints
(assuming that $w=v+1$). For instance, if shipment $i$ is processed
first, crane $v$ must move to bay $1$ before shipment $j$ can be
processed. Finally, the set of interferences can be defined by
\[
\Theta= \{(i,j,v,w) \in C^2 \times QC^2 \mid i<j \ \& \ \Delta_{i,j}^{v,w}>0 \}.
\]
The dummy (initial and last) shipments are only used in the
MIP model.

\begin{table}[!t]
\caption{The Parameters of the IPCTP.}
\label{table-parameter}
\vspace{0.2cm}
\begin{tabular}{ll}
\hline
$S$ & Set of berthed vessels\\
$C_{u}^{s}$ & Set of inbound shipments that belong to vessel $s \in S$\\
$C_{l}^{s}$ & Set of outbound shipments that belong to vessel $s \in S$\\
$C$ &  Set of all shipments\\
$C_{u}$ &  Set of inbound shipments\\
$C_{l}$ &  Set of outbound shipments\\
$L_{u}$ &  Set of available yard locations for inbound shipments\\
$L_{l}$ &  Set of yard locations for outbound shipments\\
$l_{i}$ &  Yard location of outbound shipment $i \in C_{l}$ \\
$L$ &  Set of all yard locations\\
$QC$ &  Set of QCs\\
$YC$ &  Set of YCs\\
$B$ &  Set of vessel bays \\
$B_{T}$ &  Total number of vessel bays\\
$QC_{T}$ &  Total number of QCs\\
$b_{i}$ &  Vessel bay position of shipment $i \in C$\\
$QC(i)$ &  Set of eligible QCs for shipment $i \in C$\\
$YC(k)$ &  The YC responsible for yard location $k \in L$\\
$w_{s}$ &  Weight (priority) of vessel $s \in S$\\
$Q_{i}$ &  QC handling time of shipment $i \in C$\\
$Y_{i}$ &  YT handling time of shipment $i \in C$\\
$tyt_{i}$ &  YT handling time of outbound shipment $i \in C_{l}$\\
$tt_{k}$ &  YT transfer time of inbound shipment to yard location $k \in L_{u}$\\
$tyc_{k,l}$ &  YC travel time between yard locations $k$ and $l$\\
$eqc_{i,j}$ &  QC travel time from shipment $i \in C$ to shipment $j \in C$\\
$eyc_{i,j}$ &  YC travel time from yard location $i \in L$ to yard location $j \in L$\\
$s_{QC}$ &  Travel time for unit distance of equipment $QC$\\
$\delta$ &  Safety distance between two $QC$s\\
$\delta_{v,w}$ &  Smallest allowed difference between bay positions of quay cranes $v$ and $w$ \\
$\Delta_{i,j}^{v,w}$ &  Minimum time between the starting times of shipments $i$ and $j$ \\
                    & when processed by cranes $v$ and $w$ \\
$\Theta$ &  Set of all combinations of shipments and QCs with potential interferences \\
$0$ &  Dummy initial shipment\\
$N$ &  Dummy last shipment\\
$C^0$ &  Set of all shipments including dummy initial shipment $C \cup \{0\}$ \\
$C^N$ &  Set of all shipments including dummy last shipment $C \cup \{N\}$ \\
$M$ &  A sufficiently large constant integer \\
\hline
\end{tabular}
\end{table}

\section{Literature Review}
\label{section-related-work}

Port container terminal operations have received significant attention
and many studies are dedicated to the sub-problems described earlier:
See \cite{RePEc:eee:ejores:v:244:y:2015:i:3:p:675-689} for a classification of these
subproblems. Recent work often consider the integration of two or
three problems but very few papers propose formulations covering all
the sub-problems jointly. Some papers give mathematical formulations
of integrated problems but only use heuristic approaches given the
computational complexity of solving the models. This section reviews
recent publications addressing the integrated problems and highlights
their contributions.

Chen et al. \cite{Chen2007} propose a hybrid flowshop scheduling
problem (HFSP) to schedule QCs, YTs, and YCs jointly.  Both outbound
and inbound operations are considered, but outbound operations only
start after all inbound operations are complete. In the proposed
mathematical model, each stage of the flowshop has unrelated multiple
parallel machines and a tabu-search algorithm is used to address the
computational complexity. Zheng et al. \cite{5421359} study the
scheduling of QCs and YCs, together with the yard storage and vessel
stowage plans. The authors consider an automated container handling
system, in which twin 40’ QCs and a railed container handling system
is used.  A rough yard allocation plan is maintained to indicate which
blocks are available for storing the outbound containers from each
bay. No mathematical model is provided, and the yard allocation, vessel
stowage, and equipment scheduling is performed using a rule-based
heuristic.

Xue et al. \cite{Xue2013} propose a mixed integer programming (MIP) model for
integrating the yard location assignment for inbound containers, quay
crane scheduling, and yard truck scheduling. Non-crossing constraints
and safety distances between QCs are ignored, and the assignment
of QCs and YTs are predetermined. The yard location assignment
considers block assignments instead of container slots. The resulting
model cannot be solved for even medium-sized problems. Instead, a
two-stage heuristic algorithm is employed, combining an ant colony
optimization algorithm, a greedy algorithm, and local search.

Chen et al. \cite{CHEN2013142} consider the integration of quay crane
scheduling, yard crane scheduling, and yard truck transportation. The
problem is formulated as a constraint-programming model that includes
both equipment assignment and scheduling. However, non-crossing
constraints and safety margins are ignored. The authors state that
large-scale instances are computationally intractable for constraint
programming and that even small-scale instances are too
time-consuming.  A three-stage heuristic algorithm is solved
iteratively to obtain solutions for large-scale problems with up to
500 containers.

Wu et al. \cite{WU201313} study the scheduling of different types of
equipment together with the storage strategy in order to optimize yard
operations. Only loading operations for outbound containers are
considered, and the tasks assigned to each QC and their processing
sequence are assumed to be known. The authors formulate models to
schedule the YCs and automated guided vehicles (AGV), and use a genetic
algorithm to solve large-scale problems.

Homayouni et al. \cite{Homayouni2012} study the integrated scheduling
of cranes, vehicles, and storage platforms at automated container
terminals. They consider a split-platform/retrieval system (SP-AS/RS),
which includes AGVs and handling platforms for
storing containers efficiently and providing quick access. A
mathematical model of the same problem is proposed in
\cite{HOMAYOUNI2014545}. In these studies, both outbound and inbound
operations are considered. The origin and destination points of the
containers are assumed to be predetermined and, in addition, empty for
inbound containers. The earlier study proposes a simulated annealing
(SA) algorithm to solve the problem, whereas the latter proposes a
genetic algorithm (GA) outperforming SA under the same assumptions.

Lu and Le \cite{LU2014209} propose an integrated optimization of container
terminal scheduling, including YTs and YCs. The
authors consider uncertainty factors such as YT travel speed,
YC speed, and unit time of the YC operations. The
assignment of YTs and YCs are not considered, and pre-assignments are
assumed. The objective is to minimize the operation time of YCs in
coordination with the YTs and QCs. The authors use a simulation of the
real terminal operation environment to capture uncertainties.  The
authors also formulate a mathematical model and propose a particle
swarm optimization (PSO) algorithm. As a future study, they indicate
that the scheduling for simultaneous outbound and inbound operations
should be considered for terminals adopting parallel operations.

Finally, a few additional studies
\cite{CHEN200740,Homayouni2013,KAVESHGAR2015168,LAU2008665,5223935,NIU2016284,TANG2014978,XIN2015377,XIN2014214}
integrate different sub-problems, highlighting the increasing
attention given to integrated solutions. They propose a wide range of
heuristic or meta-heuristic algorithms; e.g., genetic algorithm, tabu
search, particle swarm optimization, and rule-based heuristic methods.

Although most papers in container port operations focus on individual
problems, recent developments have emphasized the need and potential
for coordinating these interdependent operations. This paper pushes
the state-of-the-art further by optimizing all operations holistically
and demonstrating that constraint programming is a strong vehicle to
address this integrated problem.

\section{The MIP Model}
\label{section-MIP}

\begin{model}[!t]
\caption{The MIP Model for the IPCTP: Decision Variables}
\label{model:MIP-DV}
\begin{subequations}
\vspace{-0.2cm}
\begin{align}
& \mbox{\bf Variables} \nonumber \\
& x_{i,k} \in \{0,1\}: \mbox{inbound shipment } i \mbox{ is assigned to yard location } k \nonumber \\
& z_{i,j}^{q} \in \{0,1\}: \mbox{inbound shipment }j \mbox{ is handled immediately after shipment }i \mbox{ by QC }q \nonumber \\
& qz_{i,j} \in \{0,1\}: \mbox{inbound shipment }j \mbox{ is handled after shipment }i \mbox{ by QC } \nonumber \\
& v_{i,j}^{c} \in \{0,1\}: \mbox{ inbound shipment }j \mbox{ is handled immediately after shipment }i \mbox{ by YC }c \nonumber \\
& sqc_{i} \geq 0: \mbox{start time of shipment }i \mbox{ by its QC} \nonumber \\
& syc_{i} \geq 0 : \mbox{start time of shipment }i \mbox{ by its YC} \nonumber \\
& t_{i} \geq 0: \mbox{travel time of YT for inbound shipment } i \mbox{ to assigned yard location }k \nonumber \\
& sy_{i,j} \geq 0: \mbox{travel time of YC from location } i \mbox{ to location } j \nonumber \\
& Cmax_{s}: \mbox{time of the last handled container at vessel }s \nonumber
\end{align}
\end{subequations}
\end{model}

\begin{model}[!t]
\caption{The MIP Model for the IPCTP: Objective and Constraints}
\label{model:MIP}
\begin{subequations}
\vspace{-0.2cm}
\begin{align}
& \mbox{\bf Objective} \nonumber \\
&  \mbox{minimize } \textstyle\sum_{s \in S} Cmax_{s} \\
& \mbox{\bf Constraints} \nonumber \\
&  Cmax_{s} \geq w_{s}\left(sqc_{i}+Q_{i}\right)\quad \forall s \in S, \forall i \in C_{l}^{s}\enspace \\
&  Cmax_{s} \geq w_{s}\left(syc_{i}+Y_{i}\right)\quad \forall s \in S, \forall i \in C_{u}^{s} \\
&  \textstyle\sum_{i \in C_{u}} x_{i,k} \le 1\quad \forall k \in L_{u} \\
&  \textstyle\sum_{k \in L_{u}} x_{i,k} = 1\quad \forall i \in C_{u} \\
&  \textstyle\sum_{j \in C^{N}} z_{0,j}^{q} = 1\quad \forall q \in QC \\
&  \textstyle\sum_{j \in C^{N}} v_{0,j}^{c} = 1\quad \forall c \in YC \\
&  \textstyle\sum_{j \in C^{0}} z_{i,N}^{q} = 1\quad \forall q \in QC \\
&  \textstyle\sum_{j \in C^{0}} v_{i,N}^{c} = 1\quad \forall c \in YC \\
&  \textstyle\sum_{q \in QC(i)} \sum_{j \in C^{N}} z_{i,j}^{q} = 1\quad \forall i \in C, i \ne j \\
&  \textstyle\sum_{j \in C^{N}} v_{i,j}^{YC(k)} = x_{i,k}\quad \forall k \in L_{u},\forall i \in C_{u}, i \ne j \\
&  \textstyle\sum_{j \in C^{N}} v_{i,j}^{YC(k)} =1\quad \forall k \in L_{l},\forall i \in C_{l}, i \ne j \\
&  \textstyle\sum_{j \in C^{0}} z_{j,i}^{q}-\sum_{j \in C^{N}} z_{i,j}^{q} =0\quad \forall i \in C,\forall q \in QC \\
&  \textstyle\sum_{j \in C^{0}} v_{j,i}^{c}-\sum_{j \in C^{N}} v_{i,j}^{c} =0\quad \forall i \in C,\forall c \in YC \\
&  \textstyle t_{i}=\sum_{k \in L_{u}} \left(tt_{k}*x_{i,k}\right)\quad \forall i \in C_{u} \\
&  \textstyle sy_{i,j}=\sum_{m \in L_{u}} \left(tyc_{m,l_i}*x_{i,m}\right)\quad \forall i \in C_{u}, \forall j \in C_{l} \\
&  \textstyle sy_{i,j}=\sum_{m \in L_{u}}\sum_{l \in L_{u}} \left(tyc_{m,l}*x_{i,m}*x_{j,l}\right)\quad \forall i,j \in C_{u} \\
&  \textstyle sy_{i,j}=\sum_{m \in L_{u}} \left(tyc_{l_i,m}*x_{i,m}\right)\quad \forall i \in C_{l}, \forall j \in C_{u} \\
&  \textstyle sqc_{j}+M\left(1-z_{i,j}^{q}\right)\geq sqc_{i}+Q_{i}+eqc_{i,j}\quad \forall i,j \in C, \forall q \in QC \\
&  \textstyle syc_{j}+M\left(1-v_{i,j}^{c}\right)\geq syc_{i}+Y_{i}+sy_{i,j}\quad \forall i \in C_{u},\forall j \in C, \forall c \in YC \\
&  \textstyle syc_{j}+M\left(1-v_{i,j}^{c}\right)\geq syc_{i}+Y_{i}+sy_{i,j}\quad \forall i \in C_{l},\forall j \in C_{u}, \forall c \in YC \\
&  \textstyle syc_{j}+M\left(1-v_{i,j}^{c}\right)\geq syc_{i}+Y_{i}+eyc_{i,j}\quad \forall i,j \in C_{l}, \forall c \in YC \\
&  \textstyle sqc_{i}\geq syc_{i}+Y_{i}+tyt_{i}\quad \forall i \in C_{l} \\
&  \textstyle syc_{i}\geq sqc_{i}+Q_{i}+t_{i}\quad \forall i \in C_{u} \\
&  \textstyle sqc_{i}+Q_{i}-sqc_{j}\le M\left(1-qz_{i,j}\right)\quad \forall i,j \in C \\
&  \textstyle \sum_{u \in C^{0}} z_{u,i}^{v}+\sum_{u \in C^{0}} z_{u,j}^{w}\le 1+qz_{i,j}+qz_{j,i}\quad \forall \left(i,j,v,w\right) \in \Theta \\
&  \textstyle sqc_{i}+Q_{i}+\Delta_{i,j}^{v,w}-sqc_{j}\le M\left(3-qz_{i,j}-\sum_{u \in C^{0}}z_{u,i}^{v}-\sum_{u \in C^{0}}z_{u,j}^{w}\right)\quad \forall \left(i,j,v,w\right) \in \Theta
\end{align}
\end{subequations}
\end{model}

The MIP decision variables are presented in Model \ref{model:MIP-DV},
while the objective function and the constraints are given in Model
\ref{model:MIP}. They use the formulation of QC interference
constraints from \cite{Bierwirth2009}.

The first set of MIP variables are binary: They determine the yard
location of every inbound shipment and the (immediate) successor
relationships on the cranes. The remaining variables are essentially
devoted to the start times and the travel times of the shipments.

The objective function (2-01) minimizes the maximum weighted
completion time of each vessel. Constraints (2-02--2-03) compute the
weighted completion time of each vessel. Inbound shipments start their
operations at a QC and finish at a YC, whereas outbound shipments
follow the reverse order. Constraint (2-04) expresses that each
available storage block stores at most one inbound
shipment. Constraint (2-05) ensures that each inbound shipment is
assigned an available storage block. All the containers of a shipment
are assigned to the same block.  Constraints (2-06--2-07) assigns the
first (dummy) shipments to each QC and YC and Constraints (2-08--2-09)
do the same for the last (dummy) shipments. Constraint (2-10) states
that every shipment is handled by exactly one eligible QC.
Constraints (2-11--2-12) ensures that each shipment is handled by a
single YC.  In Constraint (2-12), yard blocks are known at the
beginning of the planning horizon for outbound shipments: They are
thus directly assigned to the dedicated YCs. Constraints (2-13--2-14)
guarantee that the shipments are handled in well-defined sequences by
each handling equipment (QC and YC). Constraint (2-15) defines the YT
transportation times for the inbounds shipments.  Constraints
(2-16--2-18) specify the empty travel times of the YCs according to
yard block assignments of the shipments. Constraints (2-19--2-22)
specify the relationship between the start times of two consecutive
shipments processed by the same handling equipment.  Constraints
(2-23--2-24) are the precedence constraints for each shipment,
which again differ for inbound and outbound shipments.  Constraint
(2-25) ensures that, if shipment $i$ precedes shipment $j$ on a QC,
shipment $j$ cannot start its operation on that QC until shipment $i$
finishes. Constraint (2-26) guarantees that shipments that potentially
interfere are not allowed to be processed at the same time on any
QC. Constraint (2-27) imposes a minimum temporal distance between the
processing of such shipments, which corresponds to the time taken by
the QC to move to a safe location.

There are nonlinear terms in constraint (2-17), which computes the
empty travel time of a YC, i.e., when it travels between the
destinations of two inbound shipments. These terms can be
linearized by introducing new binary variables of the form $\theta_{i,k,j,l}$
to denote whether inbound shipments $i,j\in C_{u}$ are assigned to
yards locations $k,l\in L_{u}$. The constraints then become:
\begin{align}
& \textstyle sy_{i,j}=\sum_{k \in L_{u}}\sum_{l \in L_{u}}tyc_{k,l}\theta_{i,k,j,l}\quad \forall i,j \in C_{u} \nonumber \\
& \textstyle \theta_{i,k,j,l}\geq x_{i,k}+x_{j,l}-1, \forall i,j \in C_{u}\quad \forall k,l \in L_{u},i\ne j,k\ne l \nonumber \\
& \textstyle 2-\left(x_{i,k}+x_{j,l}\right)\le 2\left(1-\theta_{i,k,j,l}\right)\quad \forall i,j \in C_{u},\forall k,l \in L_{u},i\ne j,k\ne l \nonumber
\end{align}

\section{The Constraint Programming Model}
\label{section-CP}

\begin{model}[!t]
\caption{The CP Model for the IPCTP}
\label{model:CP-DV}
\begin{subequations}
\vspace{-0.2cm}
\begin{align}
& \mbox{\bf Variables} \nonumber \\
& qc_{i}: \mbox{Interval variable for the QC handling of shipment }i \nonumber \\
& yt_{i}: \mbox{Interval variable for the YT handling of inbound shipment }i \nonumber \\
& aqc_{i,j}: \mbox{Optional interval variable for shipment }i \mbox{ on QC }j \mbox{ with duration }Q_{i} \nonumber \\
& ayc_{i,k}: \mbox{Optional interval variable for shipment }i \mbox{ on YC at yard location }k \mbox{ with duration }Y_{i} \nonumber \\
& qcs_{j}: \mbox{Sequence variable for QC } j \mbox{ over } \{ aqc_{i,j} \mid i \in C \} \nonumber \\
& ycs_{j}: \mbox{Sequence variable for YC }j \mbox{ over } \{ ayc_{i,k} \mid i \in C \wedge YC(k) = j\} \nonumber \\
& interfere_{i,v,j,w}: \mbox{Sequence variable over } \{aqc_{i,v},aqc_{j,w}\} \nonumber \\
%
& \mbox{\bf Objective} \nonumber \\
& \mbox{minimize } \textstyle \sum_{s \in S} w_{s} \left(max\left(\max_{i\in C_{u}^{s}}\textsc{endOf}\left(yt_{i}\right),\max_{j\in C_{l}^{s}}\textsc{endOf}\left(qc_{j}\right)\right)\right)  \\
& \mbox{\bf Constraints} \nonumber \\
& \textstyle \textsc{alternative}\left(qc_{i}, \mbox {all}\left(\mbox{$j$ in } QC\left(i\right)\right)aqc_{i,j}\right)\quad \forall i\in C\enspace \\
& \textstyle \textsc{alternative}\left(yt_{i}, \mbox {all}\left(\mbox{$k$ in } L_{u}\right)ayc_{i,k}\right)\quad \forall i\in C_{u} \\
& \textstyle \sum_{i \in C_{u}}\textsc{presenceOf}\left(ayc_{i,k}\right)\le 1\quad \forall k\in L_{u} \\
& \textstyle \textsc{presenceOf}\left(ayc_{i,l_{i}}\right)=1\quad \forall i\in C_{l} \\
& \textstyle \textsc{noOverlap}\left(ycs_{m},eyc_{i,j}\right)\quad \forall m\in YC \\
& \textstyle \textsc{noOverlap}\left(qcs_{m},eqc_{i,j}\right)\quad \forall m\in QC \\
& \textstyle \textsc{endBeforeStart}\left(aqc_{i,n},ayc_{i,k},tt_{k}\right)\quad \forall i\in C_{u}, k\in L_{u}, n\in QC_{i} \\
& \textstyle \textsc{endBeforeStart}\left(ayc_{i,l_{i}},aqc_{i,n},tyt_{i}\right)\quad \forall i\in C_{l}, n\in QC_{i} \\
& \textstyle \textsc{noOverlap}\left(interfere_{i,v,j,w},\Delta_{i,j}^{v,w}\right)\ \forall i,j\in C, v\in QC_{i}, w\in QC_{j}:\Delta_{i,j}^{v,w}>0
\end{align}
\end{subequations}
\end{model}

The CP model is presented in Model \ref{model:CP-DV} using the OPL API
of CP Optimizer. It uses interval variables for representing the QC
handling of all shipments and the YC handling of inbound shipments. In addition, a range of optional interval variables are used to represent the
handling of shipment $i$ on QC $j$, and the handling of shipment $i$ at
yard location $k$. The model also declares a number of sequence variables
associated with each QC and YC: Each sequence constraint collects all
the optional interval variables associated with a specific
crane. Finally, the model declares a number of sequences for optional
interval variables that may interfere.\footnote{Not all such sequences
  are useful but we declare them for simplicity.}

The CP model minimizes the weighted completion time of each vessel by
computing the maximum end date of the yard cranes (inbound shipments)
and for the quay crane (outbound shipments). Alternative constraints
(3-02) ensure that the QC processing of a shipment is performed by
exactly one QC.  Alternative constraints (3-03) enforce that each
inbound shipment is allocated to exactly one yard location, and hence
one yard crane. Constraints (3-04) state that at most one shipment can
be allocated to each yard location. Constraints (3-05) fix the yard
location (and hence the yard crane) of outbound shipments. Cranes are
disjunctive resources and can execute only one task at a time, which
is expressed by the {\sc noOverlap} constraints (3-06--3-07) over the
sequence variables associated with the cranes. These constraints also
enforce the transition times between successive operations, capturing
the empty travel times between yard locations (constraints 3-06) and
bay locations (constraints 3-07). Constraints (3-08) impose the
precedence constraints between the QC and YC tasks of inbound
shipments, while adding the travel time to move the shipment from its
bay to its chosen yard location. Constraints (3-09) impose the
precedence constraints between the YC and QC operations of
outbound shipments, adding the travel time from the fixed yard
location to the fixed bay of the shipment. Interference constraints
for the QCs are imposed by constraints (3-10). These constraints state
that, if there is a conflict between two shipments and their QCs, then
the two shipments cannot overlap in time, and their executions must be
separated by a minimum time. This is
expressed by {\sc noOverlap} constraints over sequences consisting of
the pairs of optional variables associated with these tasks.

\section{Experimental Results}
\label{section-experiments}

The MIP and CP models were written in OPL and run on the IBM ILOG
CPLEX 12.7.1 software suite. The results were obtained on an Intel
Core i7-5500U CPU 2.40 GHz computer.

\subsection{Data Generation}

The test cases were generated in accordance with earlier work, while
capturing the operations of an actual container terminal. These are
the first instances of this type, since the IPCTP has not been
considered before in the literature. Travel and processing times in
the test cases model those in the actual terminal. Different instances
have slightly different times, as will become clear.

\begin{figure}[!t]
\small
	\centering
	\includegraphics[width=0.5\linewidth]{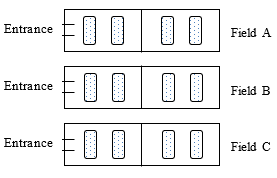}
	\caption{Layout of the Yard Side.}
	\label{figure-3}
\end{figure}

Figure \ref{figure-3} depicts the layout of the yard side considered
in the experiments, which also models the actual terminal. The yard
side is divided into 3 separate fields denoted by A, B, and C. Each
field has two location areas and a single YC is responsible for each
location area, giving a total of 6 yard cranes. In each location
area, there are 2 yard block groups, shown as the dotted region, and
traveling between them takes one unit of time. Field C is the nearest
to the quay side, and there is a hill from field C to field A. The
transportation times for YTs are generated according to these
distances. YTs can enter and exit each field from the entrance shown
in the figure, so the transfer times between the vessels and the yard
blocks close to the entrance take less time. YT transfer times are
generated between [5,10] considering the position of the yard
blocks. At the quay side, the travel of a QC between consecutive
vessel bay locations takes 3 units of time.

The processing times of the cranes for a single container are
generated uniformly in [2,5] for YCs and [2,4] for QCs. The safety
margin for the QCs is set to 1 vessel bay.  The IPCTP is expressed in
terms of shipments and the number of containers in each shipment is
uniformly distributed between [4,40].

The experiments evaluate the impact of the number of shipments, the
number of vessel bays, the inbound-outbound shipment ratio, and the
number of available yard locations for inbound shipments. The number
of shipments varies between 5 and 25, by increments of 5. The
instances can thus contain up to 1,000 containers. The number of
vessel bays are taken in $\{4,6,8\}$. The number of QCs depends on the
vessel bays due to the QC restrictions: There are half as many QCs as
there are vessel bays. The inbound-outbound shipment ratios are 20\%
and 50\%, representing the fraction of inbound shipments over the
outbound shipments. Finally, the number of available yard locations
(U-L ratio) is computed from the number of inbound shipments: There
are 2 to 3 times more yard locations than inbound shipments. For each
configuration of the parameters, 5 random instances were generated.

\subsection{Computational Results and Analysis}

\paragraph{The Results}

The results are given in Tables \ref{Table-20} and \ref{Table-50} for
each configuration, for a total of 300 instances. Table \ref{Table-20}
reports the results for 20\% inbound-outbound ratio, and Table
\ref{Table-50} for 50\%.  In the tables, each configuration is
specified in terms of the U-L ratio, the number of bays and the number
of shipments (Shp.). The average number of containers (Cnt.) in each
shipment is also presented. For each such configuration, the tables
report the average objective value and the average CPU time for its
five instances. The CPU time is limited to an hour (3,600
seconds). The MIP solver did not always find feasible solutions within
an hour for some or all five instances of a configuration. Note that
this may result in an average objective value that is lower for the
MIP model than the CP model, even when CP solves all instances
optimally, since the MIP model may not find a feasible solution to an
instance with a high optimal value. These cases are flagged by
superscripts of the form $x/y$, where $x$ is the number of infeasible
and $y$ is the number of suboptimal solutions in that average. The
superscripts for CP indicate the number of suboptimal solutions. An
entry 'NA' in the table means that the MIP cannot find a feasible
solution to any of the five instances. For the MIP, the tables also
report the optimality gap on termination, i.e., the gap in percentage
between the best lower and upper bounds.


For CP, the experiments were also run with a CPU limit of 600
seconds. The relative percentage deviations (RPD\%) from the 1-hour
runs are listed to assess CP's ability to find high-quality solutions
quickly. The RPD is computed as follows:
\[
RPD\% = \dfrac{\left(\mbox{Obj. in 600 sec.} - \mbox{Obj. in 3600 sec.}\right)*100}{\left(\mbox{Obj. in 3600 sec.}\right)}.
\]

\begin{table}[!t]
\small
\centering
\caption{Results for Import-Export Rate 20\%}
\label{Table-20}
\vspace{0.5cm}
\begin{tabular}{ccrr|r|r|r|r|r|r|}
\cline{5-10}
\multicolumn{1}{l}{}                      & \multicolumn{1}{l}{}                                                      & \multicolumn{1}{l}{}                                                          & \multicolumn{1}{l|}{}                                                                & \multicolumn{3}{c|}{\textbf{MIP}}                                                                                                          & \multicolumn{3}{c|}{\textbf{CP}}                                                                                                            \\ \hline
\multicolumn{1}{|c|}{\begin{tabular}[c]{@{}c@{}}U-L\\ Ratio\end{tabular}} & \multicolumn{1}{c|}{\begin{tabular}[c]{@{}c@{}}\# of\\ Bays\end{tabular}} & \multicolumn{1}{c|}{\begin{tabular}[c]{@{}c@{}}\# of\\ Shp.\end{tabular}} & \multicolumn{1}{c|}{\begin{tabular}[c]{@{}c@{}}Avg. \#\\ of Cnt.\end{tabular}} & \multicolumn{1}{c|}{Obj.} & \multicolumn{1}{c|}{\begin{tabular}[c]{@{}c@{}}CPU\\ (sec.)\end{tabular}} & \multicolumn{1}{c|}{GAP\%} & \multicolumn{1}{c|}{Obj.} & \multicolumn{1}{c|}{\begin{tabular}[c]{@{}c@{}}CPU\\ (sec.)\end{tabular}} & \multicolumn{1}{c|}{RPD\%} \\
\hline
\multicolumn{1}{|c|}{\multirow{15}{*}{2}} & \multicolumn{1}{c|}{\multirow{5}{*}{4}}                                   & \multicolumn{1}{r|}{5}                                                        & 67.8                                                                                 & 301.80                    & 0.06                                                                      & 0.00                       & 301.80                    & 0.27                                                                      & 0.00                       \\ \cline{3-10}
\multicolumn{1}{|c|}{}                    & \multicolumn{1}{c|}{}                                                     & \multicolumn{1}{r|}{10}                                                       & 88.8                                                                                 & $548.20^{0/3}$                    & 3032.78                                                                   & 0.17                       & 548.20                    & 5.45                                                                      & 0.00                       \\ \cline{3-10}
\multicolumn{1}{|c|}{}                    & \multicolumn{1}{c|}{}                                                     & \multicolumn{1}{r|}{15}                                                       & 99.8                                                                                 & $751.20^{0/5}$                    & 3604.88                                                                   & 0.56                       & 748.80                    & 32.78                                                                     & 0.00                       \\ \cline{3-10}
\multicolumn{1}{|c|}{}                    & \multicolumn{1}{c|}{}                                                     & \multicolumn{1}{r|}{20}                                                       & 117                                                                                  & $829.40^{0/5}$                    & 3600.38                                                                   & 0.62                       & $820.00^{1}$                    & 775.74                                                                    & 0.00                       \\ \cline{3-10}
\multicolumn{1}{|c|}{}                    & \multicolumn{1}{c|}{}                                                     & \multicolumn{1}{r|}{25}                                                       & 139.4                                                                                & $1032.75^{1/4}$                   & 3600.25                                                                   & 0.66                       & $999.00^{2}$                    & 1700.54                                                                   & 0.06                       \\ \cline{2-10}
\multicolumn{1}{|c|}{}                    & \multicolumn{1}{c|}{\multirow{5}{*}{6}}                                   & \multicolumn{1}{r|}{5}                                                        & 157.2                                                                                & 242.00                    & 0.06                                                                      & 0.00                       & 242.00                    & 0.34                                                                      & 0.00                       \\ \cline{3-10}
\multicolumn{1}{|c|}{}                    & \multicolumn{1}{c|}{}                                                     & \multicolumn{1}{r|}{10}                                                       & 168.2                                                                                & 389.60                    & 68.51                                                                     & 0.00                       & 389.60                    & 6.60                                                                      & 0.00                       \\ \cline{3-10}
\multicolumn{1}{|c|}{}                    & \multicolumn{1}{c|}{}                                                     & \multicolumn{1}{r|}{15}                                                       & 193.4                                                                                & $560.20^{0/5}$                    & 3607.59                                                                   & 0.42                       & 553.60                    & 47.90                                                                     & 0.00                       \\ \cline{3-10}
\multicolumn{1}{|c|}{}                    & \multicolumn{1}{c|}{}                                                     & \multicolumn{1}{r|}{20}                                                       & 207.8                                                                                & $634.67^{2/3}$                    & 3600.99                                                                   & 0.50                       & 647.40                    & 222.27                                                                    & 0.00                       \\ \cline{3-10}
\multicolumn{1}{|c|}{}                    & \multicolumn{1}{c|}{}                                                     & \multicolumn{1}{r|}{25}                                                       & 233.8                                                                                & $916.00^{4/1}$                    & 3608.07                                                                   & 0.62                       & 754.20                    & 755.66                                                                    & 0.15                       \\ \cline{2-10}
\multicolumn{1}{|c|}{}                    & \multicolumn{1}{c|}{\multirow{5}{*}{8}}                                   & \multicolumn{1}{r|}{5}                                                        & 246.8                                                                                & 584.00                    & 0.13                                                                      & 0.00                       & 583.90                    & 1.18                                                                      & 0.00                       \\ \cline{3-10}
\multicolumn{1}{|c|}{}                    & \multicolumn{1}{c|}{}                                                     & \multicolumn{1}{r|}{10}                                                       & 254.8                                                                                & 640.60                    & 23.88                                                                     & 0.00                       & 640.60                    & 23.88                                                                     & 0.00                       \\ \cline{3-10}
\multicolumn{1}{|c|}{}                    & \multicolumn{1}{c|}{}                                                     & \multicolumn{1}{r|}{15}                                                       & 279.4                                                                                & $994.00^{0/5}$                    & 3607.15                                                                   & 0.32                       & 952.10                    & 160.57                                                                    & 0.00                       \\ \cline{3-10}
\multicolumn{1}{|c|}{}                    & \multicolumn{1}{c|}{}                                                     & \multicolumn{1}{r|}{20}                                                       & 306.2                                                                                & $1261.00^{0/5}$                   & 3600.63                                                                   & 0.45                       & 1161.90                   & 835.41                                                                    & 0.40                       \\ \cline{3-10}
\multicolumn{1}{|c|}{}                    & \multicolumn{1}{c|}{}                                                     & \multicolumn{1}{r|}{25}                                                       & 325                                                                                  & NA                        & NA                                                                        & NA                         & 1343.10                   & 1855.45                                                                   & 3385.58                    \\ \hline
\multicolumn{1}{|c|}{\multirow{15}{*}{3}} & \multicolumn{1}{c|}{\multirow{5}{*}{4}}                                   & \multicolumn{1}{r|}{5}                                                        & 338                                                                                  & 244.40                    & 0.09                                                                      & 0.00                       & 244.40                    & 0.33                                                                      & 0.00                       \\ \cline{3-10}
\multicolumn{1}{|c|}{}                    & \multicolumn{1}{c|}{}                                                     & \multicolumn{1}{r|}{10}                                                       & 358.6                                                                                & $434.00^{0/2}$                    & 1488.63                                                                   & 0.10                       & 434.00                    & 5.10                                                                      & 0.00                       \\ \cline{3-10}
\multicolumn{1}{|c|}{}                    & \multicolumn{1}{c|}{}                                                     & \multicolumn{1}{r|}{15}                                                       & 378.2                                                                                & $642.40^{0/5}$                    & 3604.06                                                                   & 0.49                       & 641.20                    & 38.15                                                                     & 0.00                       \\ \cline{3-10}
\multicolumn{1}{|c|}{}                    & \multicolumn{1}{c|}{}                                                     & \multicolumn{1}{r|}{20}                                                       & 390.4                                                                                & $863.80^{0/5}$                    & 3603.24                                                                   & 0.64                       & $860.20^{1}$                    & 840.29                                                                    & 0.00                       \\ \cline{3-10}
\multicolumn{1}{|c|}{}                    & \multicolumn{1}{c|}{}                                                     & \multicolumn{1}{r|}{25}                                                       & 402.6                                                                                & $920.00^{3/2}$                    & 3625.08                                                                   & 0.65                       & $844.80^{2}$                    & 2086.27                                                                   & 0.08                       \\ \cline{2-10}
\multicolumn{1}{|c|}{}                    & \multicolumn{1}{c|}{\multirow{5}{*}{6}}                                   & \multicolumn{1}{r|}{5}                                                        & 418                                                                                  & 310.20                    & 0.22                                                                      & 0.00                       & 310.20                    & 0.45                                                                      & 0.00                       \\ \cline{3-10}
\multicolumn{1}{|c|}{}                    & \multicolumn{1}{c|}{}                                                     & \multicolumn{1}{r|}{10}                                                       & 439                                                                                  & 421.60                    & 127.63                                                                    & 0.00                       & 421.60                    & 6.96                                                                      & 0.00                       \\ \cline{3-10}
\multicolumn{1}{|c|}{}                    & \multicolumn{1}{c|}{}                                                     & \multicolumn{1}{r|}{15}                                                       & 461                                                                                  & $513.80^{0/5}$                    & 3600.55                                                                   & 0.35                       & 513.20                    & 323.83                                                                    & 0.00                       \\ \cline{3-10}
\multicolumn{1}{|c|}{}                    & \multicolumn{1}{c|}{}                                                     & \multicolumn{1}{r|}{20}                                                       & 484                                                                                  & $699.50^{3/2}$                    & 3600.68                                                                   & 0.54                       & 631.80                    & 140.84                                                                    & 0.00                       \\ \cline{3-10}
\multicolumn{1}{|c|}{}                    & \multicolumn{1}{c|}{}                                                     & \multicolumn{1}{r|}{25}                                                       & 509.4                                                                                & $659.00^{4/1}$                    & 3600.57                                                                   & 0.51                       & $856.80^{5}$                    & 3601.78                                                                   & 0.14                       \\ \cline{2-10}
\multicolumn{1}{|c|}{}                    & \multicolumn{1}{c|}{\multirow{5}{*}{8}}                                   & \multicolumn{1}{r|}{5}                                                        & 519.2                                                                                & 597.00                    & 0.13                                                                      & 0.00                       & 596.90                    & 1.42                                                                      & 0.00                       \\ \cline{3-10}
\multicolumn{1}{|c|}{}                    & \multicolumn{1}{c|}{}                                                     & \multicolumn{1}{r|}{10}                                                       & 532.8                                                                                & 815.40                    & 161.25                                                                    & 0.00                       & 815.20                    & 25.67                                                                     & 0.00                       \\ \cline{3-10}
\multicolumn{1}{|c|}{}                    & \multicolumn{1}{c|}{}                                                     & \multicolumn{1}{r|}{15}                                                       & 571.4                                                                                & $980.40^{0/5}$                    & 3600.70                                                                   & 0.30                       & 972.40                    & 232.30                                                                    & 0.00                       \\ \cline{3-10}
\multicolumn{1}{|c|}{}                    & \multicolumn{1}{c|}{}                                                     & \multicolumn{1}{r|}{20}                                                       & 591.2                                                                                & $1219.50^{3/2}$                   & 3600.16                                                                   & 0.46                       & 1214.90                   & 1475.40                                                                   & 3.34                       \\ \cline{3-10}
\multicolumn{1}{|c|}{}                    & \multicolumn{1}{c|}{}                                                     & \multicolumn{1}{r|}{25}                                                       & 611.2                                                                                & $1471.00^{4/1}$                   & 3600.27                                                                   & 0.48                       & $1449.50^{1}$                   & 2498.82                                                                   & 4390.42                    \\ \hline
\end{tabular}
\end{table}

\begin{table}[!t]
\small
\centering
\caption{Results for Import-Export Rate 50\%}
\label{Table-50}
\vspace{0.5cm}
\begin{tabular}{ccrr|r|r|r|r|r|r|}
\cline{5-10}
\multicolumn{1}{l}{}                      & \multicolumn{1}{l}{}                                                      & \multicolumn{1}{l}{}                                                          & \multicolumn{1}{l|}{}                                                                & \multicolumn{3}{c|}{\textbf{MIP}}                                                                                                          & \multicolumn{3}{c|}{\textbf{CP}}                                                                                                            \\ \hline
\multicolumn{1}{|c|}{\begin{tabular}[c]{@{}c@{}}U-L\\ Ratio\end{tabular}} & \multicolumn{1}{c|}{\begin{tabular}[c]{@{}c@{}}\# of\\ Bays\end{tabular}} & \multicolumn{1}{c|}{\begin{tabular}[c]{@{}c@{}}\# of\\ Shp.\end{tabular}} & \multicolumn{1}{c|}{\begin{tabular}[c]{@{}c@{}}Avg. \#\\ of Cnt.\end{tabular}} & \multicolumn{1}{c|}{Obj.} & \multicolumn{1}{c|}{\begin{tabular}[c]{@{}c@{}}CPU\\ (sec.)\end{tabular}} & \multicolumn{1}{c|}{GAP\%} & \multicolumn{1}{c|}{Obj.} & \multicolumn{1}{c|}{\begin{tabular}[c]{@{}c@{}}CPU\\ (sec.)\end{tabular}} & \multicolumn{1}{c|}{RPD\%} \\
\hline
\multicolumn{1}{|c|}{\multirow{15}{*}{2}} & \multicolumn{1}{c|}{\multirow{5}{*}{4}}                                   & \multicolumn{1}{r|}{5}                                                        & 67.8                                                                                 & 320.40                    & 0.21                                                                      & 0.00                       & 320.40                    & 0.47                                                                      & 0.00                       \\ \cline{3-10}
\multicolumn{1}{|c|}{}                    & \multicolumn{1}{c|}{}                                                     & \multicolumn{1}{r|}{10}                                                       & 88.8                                                                                 & $541.60^{0/5}$                    & 3600.08                                                                   & 0.37                       & 541.60                    & 15.93                                                                     & 0.00                       \\ \cline{3-10}
\multicolumn{1}{|c|}{}                    & \multicolumn{1}{c|}{}                                                     & \multicolumn{1}{r|}{15}                                                       & 99.8                                                                                 & $738.40^{0/5}$                    & 3600.97                                                                   & 0.58                       & 738.40                    & 113.76                                                                    & 0.00                       \\ \cline{3-10}
\multicolumn{1}{|c|}{}                    & \multicolumn{1}{c|}{}                                                     & \multicolumn{1}{r|}{20}                                                       & 117                                                                                  & $755.67^{2/3}$                    & 3600.31                                                                   & 0.60                       & $820.00^{1}$                    & 1110.29                                                                  & 0.00                       \\ \cline{3-10}
\multicolumn{1}{|c|}{}                    & \multicolumn{1}{c|}{}                                                     & \multicolumn{1}{r|}{25}                                                       & 139.4                                                                                & $1570.00^{4/1}$                   & 3600.63                                                                   & 0.77                       & $998.40^{2}$                    & 1989.51                                                                   & 0.06                       \\ \cline{2-10}
\multicolumn{1}{|c|}{}                    & \multicolumn{1}{c|}{\multirow{5}{*}{6}}                                   & \multicolumn{1}{r|}{5}                                                        & 157.2                                                                                & 199.00                    & 0.11                                                                      & 0.00                       & 199.00                    & 0.83                                                                      & 0.00                       \\ \cline{3-10}
\multicolumn{1}{|c|}{}                    & \multicolumn{1}{c|}{}                                                     & \multicolumn{1}{r|}{10}                                                       & 168.2                                                                                & $377.00^{0/1}$                    & 881.29                                                                    & 0.05                       & 377.00                    & 10.71                                                                     & 0.00                       \\ \cline{3-10}
\multicolumn{1}{|c|}{}                    & \multicolumn{1}{c|}{}                                                     & \multicolumn{1}{r|}{15}                                                       & 193.4                                                                                & $583.80^{0/5}$                    & 3603.98                                                                   & 0.48                       & 542.80                    & 94.11                                                                     & 0.00                       \\ \cline{3-10}
\multicolumn{1}{|c|}{}                    & \multicolumn{1}{c|}{}                                                     & \multicolumn{1}{r|}{20}                                                       & 207.8                                                                                & $829.50^{3/2}$                    & 3600.21                                                                   & 0.63                       & 648.60                    & 900.15                                                                    & 0.00                       \\ \cline{3-10}
\multicolumn{1}{|c|}{}                    & \multicolumn{1}{c|}{}                                                     & \multicolumn{1}{r|}{25}                                                       & 233.8                                                                                & $1811.00^{4/1}$                   & 3600.52                                                                   & 0.81                       & $752.20^{1}$                    & 1312.59                                                                   & 0.23                       \\ \cline{2-10}
\multicolumn{1}{|c|}{}                    & \multicolumn{1}{c|}{\multirow{5}{*}{8}}                                   & \multicolumn{1}{r|}{5}                                                        & 246.8                                                                                & 580.40                    & 0.22                                                                      & 0.00                       & 580.20                    & 2.36                                                                      & 0.00                       \\ \cline{3-10}
\multicolumn{1}{|c|}{}                    & \multicolumn{1}{c|}{}                                                     & \multicolumn{1}{r|}{10}                                                       & 254.8                                                                                & 600.00                    & 548.25                                                                    & 0.00                       & 599.80                    & 47.07                                                                     & 0.00                       \\ \cline{3-10}
\multicolumn{1}{|c|}{}                    & \multicolumn{1}{c|}{}                                                     & \multicolumn{1}{r|}{15}                                                       & 279.4                                                                                & $990.00^{1/4}$                    & 3601.38                                                                   & 0.31                       & 898.10                    & 247.72                                                                    & 0.00                       \\ \cline{3-10}
\multicolumn{1}{|c|}{}                    & \multicolumn{1}{c|}{}                                                     & \multicolumn{1}{r|}{20}                                                       & 306.2                                                                                & $1149.33^{2/3}$                   & 3600.35                                                                   & 0.38                       & 1052.90                   & 1323.69                                                                   & 3.64                       \\ \cline{3-10}
\multicolumn{1}{|c|}{}                    & \multicolumn{1}{c|}{}                                                     & \multicolumn{1}{r|}{25}                                                       & 325                                                                                  & $3846.00^{3/2}$                   & 3600.47                                                                   & 0.81                       & $1333.50^{4}$                   & 3391.34                                                                   & 4559.78                    \\ \hline
\multicolumn{1}{|c|}{\multirow{15}{*}{3}} & \multicolumn{1}{c|}{\multirow{5}{*}{4}}                                   & \multicolumn{1}{r|}{5}                                                        & 338                                                                                  & 265.00                    & 0.22                                                                      & 0.00                       & 265.00                    & 0.95                                                                      & 0.00                       \\ \cline{3-10}
\multicolumn{1}{|c|}{}                    & \multicolumn{1}{c|}{}                                                     & \multicolumn{1}{r|}{10}                                                       & 358.6                                                                                & $428.40^{0/4}$                    & 2893.00                                                                   & 0.23                       & 428.40                    & 18.07                                                                     & 0.00                       \\ \cline{3-10}
\multicolumn{1}{|c|}{}                    & \multicolumn{1}{c|}{}                                                     & \multicolumn{1}{r|}{15}                                                       & 378.2                                                                                & $653.50^{1/4}$                    & 3601.66                                                                   & 0.49                       & $641.20^{1}$                    & 813.17                                                                    & 0.00                       \\ \cline{3-10}
\multicolumn{1}{|c|}{}                    & \multicolumn{1}{c|}{}                                                     & \multicolumn{1}{r|}{20}                                                       & 390.4                                                                                & $1201.25^{1/4}$                   & 3600.46                                                                   & 0.72                       & $859.80^{1}$                    & 1230.81                                                                   & 0.00                       \\ \cline{3-10}
\multicolumn{1}{|c|}{}                    & \multicolumn{1}{c|}{}                                                     & \multicolumn{1}{r|}{25}                                                       & 402.6                                                                                & NA                        & NA                                                                        & NA                         & $841.60^{5}$                    & 3601.76                                                                   & 0.08                       \\ \cline{2-10}
\multicolumn{1}{|c|}{}                    & \multicolumn{1}{c|}{\multirow{5}{*}{6}}                                   & \multicolumn{1}{r|}{5}                                                        & 418                                                                                  & 302.40                    & 0.61                                                                      & 0.00                       & 302.40                    & 1.21                                                                      & 0.00                       \\ \cline{3-10}
\multicolumn{1}{|c|}{}                    & \multicolumn{1}{c|}{}                                                     & \multicolumn{1}{r|}{10}                                                       & 439                                                                                  & $407.80^{0/2}$                    & 1777.30                                                                   & 0.13                       & 407.80                    & 16.12                                                                     & 0.00                       \\ \cline{3-10}
\multicolumn{1}{|c|}{}                    & \multicolumn{1}{c|}{}                                                     & \multicolumn{1}{r|}{15}                                                       & 461                                                                                  & $488.00^{1/4}$                    & 3607.27                                                                   & 0.33                       & $490.60^{1}$                    & 855.09                                                                    & 0.00                       \\ \cline{3-10}
\multicolumn{1}{|c|}{}                    & \multicolumn{1}{c|}{}                                                     & \multicolumn{1}{r|}{20}                                                       & 484                                                                                  & $971.33^{2/3}$                    & 3600.57                                                                   & 0.65                       & $611.60^{1}$                    & 959.72                                                                    & 0.00                       \\ \cline{3-10}
\multicolumn{1}{|c|}{}                    & \multicolumn{1}{c|}{}                                                     & \multicolumn{1}{r|}{25}                                                       & 509.4                                                                                & NA                        & NA                                                                        & NA                         & $856.20^{5}$                    & 3602.76                                                                   & 0.44                       \\ \cline{2-10}
\multicolumn{1}{|c|}{}                    & \multicolumn{1}{c|}{\multirow{5}{*}{8}}                                   & \multicolumn{1}{r|}{5}                                                        & 519.2                                                                                & 550.80                    & 0.18                                                                      & 0.00                       & 550.50                    & 2.57                                                                      & 0.00                       \\ \cline{3-10}
\multicolumn{1}{|c|}{}                    & \multicolumn{1}{c|}{}                                                     & \multicolumn{1}{r|}{10}                                                       & 532.8                                                                                & 756.60                    & 609.18                                                                    & 0.00                       & 756.50                    & 66.82                                                                     & 0.00                       \\ \cline{3-10}
\multicolumn{1}{|c|}{}                    & \multicolumn{1}{c|}{}                                                     & \multicolumn{1}{r|}{15}                                                       & 571.4                                                                                & $1006.20^{0/5}$                    & 3600.69                                                                   & 0.33                       & 968.80                    & 427.87                                                                    & 0.00                       \\ \cline{3-10}
\multicolumn{1}{|c|}{}                    & \multicolumn{1}{c|}{}                                                     & \multicolumn{1}{r|}{20}                                                       & 591.2                                                                                & $2428.50^{3/2}$                   & 3600.54                                                                   & 0.73                       & $1146.50^{1}$                   & 2187.52                                                                   & 772.78                     \\ \cline{3-10}
\multicolumn{1}{|c|}{}                    & \multicolumn{1}{c|}{}                                                     & \multicolumn{1}{r|}{25}                                                       & 611.2                                                                                & NA                        & NA                                                                        & NA                         & $1393.30^{5}$                   & 3613.52                                                                   & 6479.76                    \\ \hline
\end{tabular}
\end{table}

\paragraph{MIP versus CP} The experimental results indicate that CP is
orders of magnitude more efficient than MIP on the IPCTP. This is
especially remarkable since this paper compares two black-box solvers.
Overall, within the time limit, the MIP model does not find feasible
solutions for 71 out of 300 instances and cannot prove optimality for
103 instances.  On all but the smallest instances, the MIP solver
cannot prove optimality for all five instances of the same
confguration.  In almost all configurations with 20 or more shipments,
the MIP solver fails to find feasible solutions on at least one of the
instances. In contrast, the CP model always find feasible solutions
and proves optimality for 260 instances out of 300 instances. CP proves
optimality on all but 12 instances in Table \ref{Table-50}, and all
but 28 in Table \ref{Table-20}. On instances where both models find
optimal solutions, the CP model is almost always 1--3 orders of
magnitude faster (except for the smallest instances). Finally, the CP
model always dominates the MIP model, in the sense that it proves
optimality every time the MIP does.

\paragraph{Short Runs} On almost all instances but the largest ones,
the CP model finds optimal, or near optimal, solutions within 10 minutes.
On the largest instances, longer CPU times are necessary to find optimal
or near-optimal solutions.

\paragraph{Sensitivity Analysis}

The sensitivity analysis is restricted to the CP model for obvious
reasons. The sensitivity of each factor is analyzed by comparing their
respective run times and objective values.  In general, the effect of
the number of bays on the solution values and on CPU times tend to be
small. In contrast, increasing the U-L ratio from 2 to 3 gives inbound
shipments more alternatives for yard locations, which typically
increases CPU times. Increasing the ratio of inbound-outbound
containers also increases problem difficulty. This is not a surprise,
since inbound shipments are more challenging, as they require a yard
location assignment, while outbound shipments have both their yard locations and
vessel bays fixed. Nevertheless, the CP model scales reasonably well
when this ratio is increased. These analyses indicate that the number
of shipments/containers is by far the most important element in
determining the computing times in the IPCTP: The other factors have a
significantly smaller impact, which is an interesting result in its
own right.

\section{Conclusion}
\label{section-conclusion}

This paper introduced the Integrated Port Container Terminal Problem
(IPCTP) which, to the best of our knowledge, integrates for the first
time, a wealth of port operations, including the yard location
assignment, the assignment of quay and yard cranes, and the scheduling
of these cranes under realistic constraints.  In particular, the IPCTP
considers empty travel time of the equipment and interference
constraints between the quay cranes. The paper proposed both an MIP
and a CP model for the IPCTP of a configuration based on an actual
container terminal, which were evaluated on a variety of
configurations regarding the number of vessel bays, the number of yard
locations, the ratio of inbound-outbound shipments, and the number of
shipments/containers. Experimental results indicate that the MIP model
can only be solved optimally for small instances and often cannot find
feasible solutions. The CP model finds optimal solutions for 87\% of
the instances and, on instances where both models can be solved
optimally, the CP model is typically 1--3 orders of magnitude faster
and proves optimality each time the MIP does.  The CP model scales
reasonably well with the number of vessel bays and yard locations, and
the ratio of inbound-outbound shipments. It also solves large
realistic instances with hundreds of containers. These results
contrast with the existing literature which typically resort to
heuristic or meta-heuristic algorithms, with no guarantee of
optimality.

Future work will be devoted to capturing a number of additional
features, including operator-based processing times, the stacking of
inbound containers using re-shuffling operations, and the scheduling
of the yard trucks.

\newpage


\begin{thebibliography}{10}
\providecommand{\url}[1]{\texttt{#1}}
\providecommand{\urlprefix}{URL }

\bibitem{RePEc:eee:ejores:v:244:y:2015:i:3:p:675-689}
Bierwirth, C., Meisel, F.: A follow-up survey of berth allocation and quay
  crane schedul-ing problems in container terminals. European Journal of
  Operational Research  244,  675--689 (2015)

\bibitem{Bierwirth2009}
Bierwirth, C., Meisel, F.: A fast heuristic for quay crane scheduling with
  interference constraints. Journal of Scheduling  12(4),  345--360 (2009)

\bibitem{Chen2007}
Chen, L., Bostel, N., Dejax, P., Cai, J., Xi, L.: A tabu search algorithm for
  the integrated scheduling problem of container handling systems in a maritime
  terminal. European Journal of Operational Research  181,  40--58 (2007)

\bibitem{CHEN200740}
Chen, L., Bostel, N., Dejax, P., Cai, J., Xi, L.: A tabu search algorithm for
  the integrated scheduling problem of container handling systems in a maritime
  terminal. European Journal of Operational Research  181(1),  40 -- 58 (2007)

\bibitem{CHEN2013142}
Chen, L., Langevin, A., Lu, Z.: Integrated scheduling of crane handling and
  truck transportation in a maritime container terminal. European Journal of
  Operational Research  225(1),  142 -- 152 (2013)

\bibitem{Homayouni2012}
Homayouni, S.M., R~Vasili, M., M~Kazemi, S., H~Tang, S.: Integrated scheduling
  of sp-as/rs and handling equipment in automated container terminals. In:
  Proceedings of International Conference on Computers and Industrial
  Engineering, CIE. vol.~2 (2012)

\bibitem{HOMAYOUNI2014545}
Homayouni, S.M., Tang, S.H., Motlagh, O.: A genetic algorithm for optimization
  of integrated scheduling of cranes, vehicles, and storage platforms at
  automated container terminals. Journal of Computational and Applied
  Mathematics  270(Supplement C),  545 -- 556 (2014)

\bibitem{Homayouni2013}
Homayouni, S., Tang, S.: Multi objective optimization of coordinated scheduling
  of cranes and vehicles at container terminals (2013)

\bibitem{KAVESHGAR2015168}
Kaveshgar, N., Huynh, N.: Integrated quay crane and yard truck scheduling for
  unloading inbound containers. International Journal of Production Economics
  159(Supplement C),  168 -- 177 (2015)

\bibitem{LAU2008665}
Lau, H.Y., Zhao, Y.: Integrated scheduling of handling equipment at automated
  container terminals. International Journal of Production Economics  112(2),
  665 -- 682 (2008)

\bibitem{5223935}
Liang, L., Lu, Z.Q., Zhou, B.H.: A heuristic algorithm for integrated
  scheduling problem of container handling system. In: 2009 International
  Conference on Computers Industrial Engineering. pp. 40--45. IEEE (2009)

\bibitem{LU2014209}
Lu, Y., Le, M.: The integrated optimization of container terminal scheduling
  with uncertain factors. Computers \& Industrial Engineering  75(Supplement
  C),  209 -- 216 (2014)

\bibitem{NAV:NAV20121}
Moccia, L., Cordeau, J.F., Gaudioso, M., Laporte, G.: A branch-and-cut
  algorithm for the quay crane scheduling problem in a container terminal.
  Naval Research Logistics (NRL)  53(1),  45--59 (2006)

\bibitem{NIU2016284}
Niu, B., Xie, T., Tan, L., Bi, Y., Wang, Z.: Swarm intelligence algorithms for
  yard truck scheduling and storage allocation problems. Neurocomputing
  188(Supplement C),  284 -- 293 (2016)

\bibitem{Sammarra2007}
Sammarra, M., Cordeau, J.F., Laporte, G., Monaco, M.F.: A tabu search heuristic
  for the quay crane scheduling problem. Journal of Scheduling  10(4),
  327--336 (2007)

\bibitem{TANG2014978}
Tang, L., Zhao, J., Liu, J.: Modeling and solution of the joint quay crane and
  truck scheduling problem. European Journal of Operational Research  236(3),
  978 -- 990 (2014)

\bibitem{Vis2003}
Vis, I., {de Koster}, R.: Transshipment of containers at a container terminal:
  An overview. European Journal of Operational Research  147,  1--16 (2003)

\bibitem{WU201313}
Wu, Y., Luo, J., Zhang, D., Dong, M.: An integrated programming model for
  storage management and vehicle scheduling at container terminals. Research in
  Transportation Economics  42(1),  13 -- 27 (2013)

\bibitem{XIN2015377}
Xin, J., Negenborn, R.R., Corman, F., Lodewijks, G.: Control of interacting
  machines in automated container terminals using a sequential planning
  approach for collision avoidance. Transportation Research Part C: Emerging
  Technologies  60(Supplement C),  377 -- 396 (2015)

\bibitem{XIN2014214}
Xin, J., Negenborn, R.R., Lodewijks, G.: Energy-aware control for automated
  container terminals using integrated flow shop scheduling and optimal
  control. Transportation Research Part C: Emerging Technologies  44(Supplement
  C),  214 -- 230 (2014)

\bibitem{Xue2013}
Xue, Z., Zhang, C., Miao, L., Lin, W.H.: An ant colony algorithm for yard truck
  scheduling and yard location assignment problems with precedence constraints.
  Journal of Systems Science and Systems Engineering  22(1),  21--37 (2013)

\bibitem{5421359}
Zheng, K., Lu, Z., Sun, X.: An effective heuristic for the integrated
  scheduling problem of automated container handling system using twin 40'
  cranes. In: 2010 Second International Conference on Computer Modeling and
  Simulation. pp. 406--410. IEEE (2010)

\end{thebibliography}

\end{document}